\documentclass{article}

\usepackage[preprint]{neurips_2024}

\usepackage[utf8]{inputenc} % allow utf-8 input
\usepackage[T1]{fontenc}    % use 8-bit T1 fonts
\usepackage{hyperref}       % hyperlinks
\usepackage{url}            % simple URL typesetting
\usepackage{booktabs}       % professional-quality tables
\usepackage{amsfonts}       % blackboard math symbols
\usepackage{nicefrac}       % compact symbols for 1/2, etc.
\usepackage{microtype}      % microtypography
\usepackage{xcolor}         % colors
\usepackage{hyperref}       % hyperlinks
\usepackage{url}            % simple URL typesetting
\usepackage{booktabs}       % professional-quality tables
\usepackage{amsfonts}       % blackboard math symbols
\usepackage{nicefrac}       % compact symbols for 1/2, etc.
\usepackage{microtype}      % microtypography
\usepackage{amsmath}
\usepackage{geometry}
\usepackage{graphicx}
\usepackage{multirow}
\usepackage[normalem]{ulem}
\useunder{\uline}{\ul}{}
\usepackage{booktabs}
\usepackage{algorithm}
\usepackage{algpseudocode}
\usepackage{float}
\usepackage{wrapfig}
\usepackage{booktabs}
\usepackage{colortbl} 
\usepackage{marvosym}
\makeatletter
\def\thanks#1{\protected@xdef\@thanks{\@thanks
        \protect\footnotetext{#1}}}
\makeatother

\title{Scalable Visual State Space Model with \\ Fractal Scanning }

\author{
  Lv Tang, HaoKe Xiao, Peng-Tao Jiang, Hao Zhang, Jinwei Chen, Bo Li\thanks{This paper is working in progress.} \\
  vivo Mobile Communication Co., Ltd\\
  \texttt{\{lvtang,xiaohaoke,pt.jiang,haozhang,jinwei.chen,libra\}@vivo.com}}
   
\begin{document}

\maketitle

\begin{abstract}
Foundational models have significantly advanced in natural language processing (NLP) and computer vision (CV), with the Transformer architecture becoming a standard backbone. However, the Transformer's quadratic complexity poses challenges for handling longer sequences and higher resolution images. To address this challenge, State Space Models (SSMs) like Mamba have emerged as efficient alternatives, initially matching Transformer performance in NLP tasks and later surpassing Vision Transformers (ViTs) in various CV tasks. To improve the performance of SSMs, one crucial aspect is effective serialization of image patches. Existing methods, relying on linear scanning curves, often fail to capture complex spatial relationships and produce repetitive patterns, leading to biases. To address these limitations, we propose using fractal scanning curves for patch serialization. Fractal curves maintain high spatial proximity and adapt to different image resolutions, avoiding redundancy and enhancing SSMs' ability to model complex patterns accurately. We validate our method in image classification, detection, and segmentation tasks, and the superior performance validates its effectiveness.
\end{abstract}

\section{Introduction} \label{introd}
Foundational models have rapidly evolved within the domains of natural language processing (NLP) and computer vision (CV), giving rise to a plethora of works~\cite{DBLP:conf/naacl/DevlinCLT19,DBLP:conf/icml/RadfordKHRGASAM21,DBLP:conf/icml/0001LXH22,DBLP:journals/jmlr/ChowdheryNDBMRBCSGSSTMRBTSPRDHPBAI23,DBLP:journals/corr/abs-2303-08774,DBLP:journals/corr/abs-2302-13971,DBLP:journals/corr/abs-2304-07193,Kirillov_2023_ICCV}, such as LLaVA~\cite{DBLP:journals/corr/abs-2304-08485} and BLIP-2~\cite{DBLP:conf/icml/0008LSH23}. The common backbone of modern foundational models is Transformer~\cite{DBLP:conf/nips/VaswaniSPUJGKP17}, a type of sequence model. Although the self-attention mechanism in Transformer ensures effective modeling of complex data across varied contexts, its quadratic complexity poses significant efficiency challenges, thus presenting challenges to the further advancement of foundational models when facing longer language lengths or larger image resolutions.

Recently, based on State Space Model (SSM), Mamba~\cite{DBLP:journals/corr/abs-2312-00752} initially achieves performance comparable to that of Transformer in NLP with linear time complexity. Subsequently, researchers have made further improvements to Mamba~\cite{DBLP:journals/corr/abs-2403-09338,DBLP:journals/corr/abs-2403-17695,DBLP:journals/corr/abs-2401-09417,DBLP:journals/corr/abs-2401-10166}, enabling it to surpass the Visual Transformer (ViT) in different CV tasks, such as image classification and semantic segmentation. These developments have inspired further exploration into SSMs to ascertain their potential as viable alternatives to ViTs. 

For fully leveraging the modeling capabilities of SSMs in CV tasks, one of the primary challenges is the effective serialization of image patches. Effective serialization can accurately represent and retain the intricate spatial relationships and structural properties inherent in images as far as possible, which are essential for SSMs to capture the representation of images. Consequently, numerous studies have designed various scanning mechanisms to optimize the serialization. For example, ViM~\cite{DBLP:journals/corr/abs-2401-09417} employs a bi-directional scanning curve that processes image patches horizontally (row by row) to capture spatial dependencies. Building upon this, VMamba~\cite{DBLP:journals/corr/abs-2401-10166} adds serialization along vertical columns (column by column), enriching the analysis of spatial relationships. LocalMamba~\cite{DBLP:journals/corr/abs-2403-09338} introduces a window mechanism that divides the image into windows, serializing image patches within each window first, and then similarly processing between windows, which helps SSMs in capturing comprehensive image features.

Although the above methods have achieved notable performance, they generally utilize linear scanning curves, such as Zigzag, which may have inherent limitations and fail to fully capture the complex spatial relationships and structural properties within images. Additionally, these linear scans tend to produce repetitive scanning cycles, leading to serialized sequences that fall into repetitive patterns. This redundancy can introduce biases in the subsequent modeling by SSMs, as the model might overfit to these repetitive patterns rather than capturing the true underlying dynamics of the image. For example, the reliance on fixed scanning patterns may limit the model's ability to adapt to dynamic changes in image resolution, further constraining its applicability across various imaging conditions.

To address these deficiencies, inspired by fractal theories~\cite{DBLP:journals/tip/GotsmanL96}, we propose employing fractal scanning curves as a more advanced method for patch serialization. Unlike linear curves, fractal curves preserve spatial proximity during the serialization process and adapt seamlessly to varying image resolutions. This adaptability ensures the preservation of the curve’s properties across different image dimensions, enabling a consistent approach to patch serialization that circumvents the limitations of repetitive scanning cycles typical of linear curves. This improved modeling capability enables SSMs to produce more accurate and contextually relevant predictions, fully harnessing the state space approach to dynamically capture and analyze complex patterns in images.

However, the application of fractal scanning curves in SSMs for image processing has identified certain limitations, particularly regarding the imperfect preservation of local adjacency and occasional disruptions in local continuity. These issues may result in the partial loss of local proximity information, which is essential for precise modeling by SSMs. To overcome this challenge, we propose a simple yet effective adjustment to the fractal curves. We implement a shift operation, adjusting the curves vertically by one pixel. This minor alteration significantly enhances local adjacency and continuity during pixel serialization. By more closely aligning the curve with the inherent spatial relationships within the image, this adjustment effectively addresses gaps and overlaps at critical junctions and complex sections of the curve, where adjacency might otherwise be compromised. In conclusion, our study contributes several key advancements, outlined as follows:

\begin{itemize}

\item We identify and address the inherent limitations of linear scanning curves traditionally used in SSMs. By analyzing their inefficiencies, we advocate for the adoption of fractal scanning curves, which are better suited to preserving the continuity of images. 

\item We further refine the fractal scanning curve approach by implementing a shift operation. This modification enhances the curve’s ability to preserve local structures and relationships within the image, thus addressing issues of imperfect adjacency and continuity disruptions commonly associated with standard fractal curves. 
 
\item We rigorously evaluate the efficacy of our innovations across three computer vision tasks: classification, segmentation, and detection. The experimental results unequivocally demonstrate enhanced performance and broad applicability of our proposed methods.

\end{itemize}

\section{Related Work}

\subsection{Vision Backbone Architecture}
In the domain of vision backbone architectures, significant advancements have been achieved through the development and refinement of several key frameworks. The existing widely used backbones contain CNN-based and ViT-based, each offering unique advantages and suited for different tasks within the field of CV. CNNs~\cite{DBLP:conf/nips/KrizhevskySH12,DBLP:journals/corr/SimonyanZ14a,DBLP:conf/cvpr/HeZRS16,DBLP:journals/corr/HowardZCKWWAA17,DBLP:conf/cvpr/RadosavovicKGHD20} have been foundational in the progress of vision models, predominantly due to their ability to capture spatial hierarchies in images. Pioneering models like AlexNet~\cite{DBLP:conf/nips/KrizhevskySH12}, VGG~\cite{DBLP:journals/corr/SimonyanZ14a}, and ResNet~\cite{DBLP:conf/cvpr/HeZRS16} have set the benchmarks in various vision tasks and continue to be pivotal in many applications. ViTs~\cite{DBLP:conf/nips/VaswaniSPUJGKP17,DBLP:conf/iclr/DosovitskiyB0WZ21,DBLP:conf/iccv/LiuL00W0LG21,DBLP:conf/eccv/TouvronCJ22} represent a paradigm shift in vision processing by applying the principles of self-attention across patches of an image, treating them akin to tokens in natural language processing. This architecture not only capitalizes on the Transformer's ability to handle long-range dependencies but also sets a robust foundation for the explosion of modern foundational models~\cite{DBLP:conf/naacl/DevlinCLT19,DBLP:conf/icml/RadfordKHRGASAM21,DBLP:conf/icml/0001LXH22,DBLP:journals/jmlr/ChowdheryNDBMRBCSGSSTMRBTSPRDHPBAI23,DBLP:journals/corr/abs-2303-08774,DBLP:journals/corr/abs-2302-13971,DBLP:journals/corr/abs-2304-07193,Kirillov_2023_ICCV,DBLP:journals/corr/abs-2304-08485,DBLP:conf/icml/0008LSH23}. However, despite their remarkable performance, ViTs face significant efficiency challenges due to their quadratic complexity, which poses obstacles to further advancements in foundational models, particularly when dealing with longer sequences or larger image resolutions.

More recently, SSMs have been proposed as an alternative backbone architecture for vision tasks~\cite{DBLP:journals/corr/abs-2403-09338,DBLP:journals/corr/abs-2403-17695,DBLP:journals/corr/abs-2401-09417,DBLP:journals/corr/abs-2401-10166,DBLP:journals/corr/abs-2403-09977}, offering linear time complexity. SSMs employ a novel approach by modeling data sequences through state transitions, which is particularly advantageous for capturing dynamic changes in images. This emerging field seeks to provide more efficient solutions compared to traditional CNNs and Transformers. A critical issue in designing SSM-based visual backbones is how to effectively serialize image patches, transitioning from 2D to 1D while preserving the structural information in the image. This ensures that the SSM can accurately capture the current feature representation of the image. In this paper, we highlight the problems with current serialization methods and introduce a fractal serialization mechanism to enhance the performance of SSM-based backbones.

\begin{figure}[!t]
    \centering
    \includegraphics[width=\linewidth]{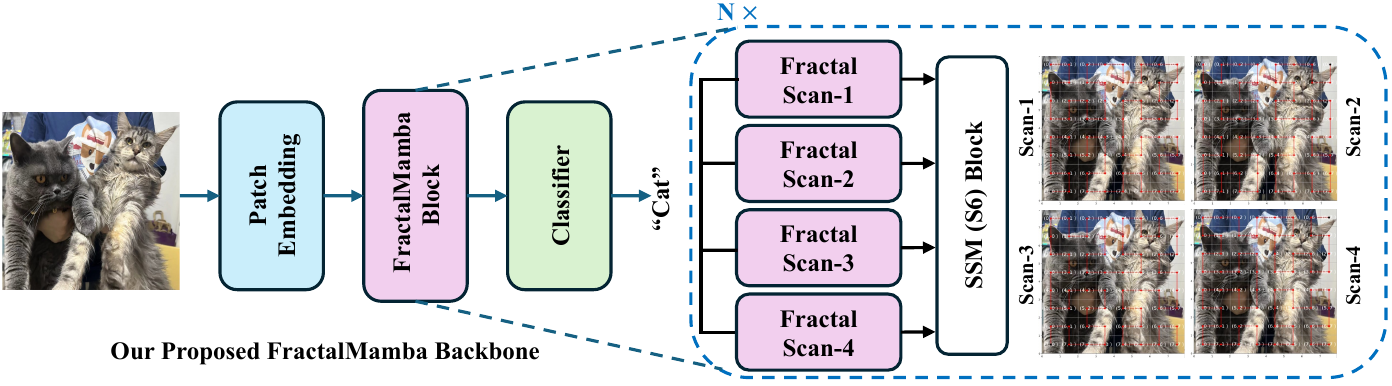}
    \caption{The architecture of our proposed FractalMamba Backbone. The FractalMamba block contains four curves in 4 directions. Taking $8 \times 8$ image patches for example, the first curve is from (0,0) to (7,0), the second curve is from (0,0) to (0,7), the third curve is from (7,7) to (7,0), and the fourth curve is from (7,7) to (0,7).}
    \label{framework}
    \vspace{-0.7cm}
\end{figure}

%希尔伯特这种方式，近期也有人做到cnn/transoformer. LOCALFORMER，transfoer没有意义。全量attention，flatten不关键。
%pos embedding之间的映射
\subsection{State Space Model}
Drawing on principles from control theory, linear state space equations have been integrated with deep learning to enhance sequential data modeling, as demonstrated in works like HiPPO~\cite{DBLP:conf/nips/GuDERR20} and LSSL~\cite{DBLP:conf/nips/GuJGSDRR21}. Recent advancements have enabled SSMs to increasingly compete with CNNs and Transformers in terms of performance. Notably, the Structured State Space Sequence Model (S4)~\cite{DBLP:conf/iclr/GuGR22} utilizes a linear state space for contextualization and has exhibited strong performance across various sequence modeling tasks, particularly with lengthy sequences. Subsequently, Mamba~\cite{DBLP:journals/corr/abs-2312-00752} has achieved a significant breakthrough with its linear-time inference and efficient training process, incorporating a selection mechanism and hardware-aware algorithms. Building on the success of Mamba, subsequent studies have explored the potential of SSMs for CV tasks~\cite{DBLP:journals/corr/abs-2403-09338,DBLP:journals/corr/abs-2403-17695,DBLP:journals/corr/abs-2401-09417,DBLP:journals/corr/abs-2401-10166}, achieving performance comparable to that of Transformers.

At the core of these endeavors is the development of various linear scanning mechanisms for serialization, as highlighted in recent surveys~\cite{xu2024survey,zhang2024survey}. However, linear scanning curves have inherent limitations and often fail to fully capture the complex spatial relationships within images. Furthermore, these linear scans can lead to repetitive scanning cycles, creating serialized sequences that exhibit redundant patterns. This redundancy can introduce biases in the subsequent modeling by SSMs, as the model may overfit to these repetitive patterns rather than capturing the true underlying dynamics of the image. To overcome these deficiencies, we propose the adoption of fractal scanning curves, which offer a more advanced approach to serialization. As shown in Fig. \ref{framework}, following works~\cite{DBLP:journals/corr/abs-2401-10166,DBLP:journals/corr/abs-2403-09338}, we use the S6 block in the FractalMamba block. Moreover, the FractalMamba block contains four curves in 4 directions.

\section{Method}
\subsection{Preliminaries}
\textbf{State Space Model.} SSM provides a robust framework for modeling physical systems, particularly Linear Time-Invariant (LTI) systems. These models excel in representing such systems through a set of first-order differential equations, effectively capturing the dynamics of the system’s state variables:
\begin{equation}
\begin{split}
h^{\prime}(t) & =\mathbf{A} h(t)+\mathbf{B} x(t), \\
y(t) & =\mathbf{C} h(t).
\end{split}
\label{first_defi}
\end{equation}
$h^{\prime}(t)$ denotes the time derivative of the state vector $h(t)$, with matrices $\mathbf{A},\mathbf{B},\mathbf{C}$ defining the relationships between the $h(t), x(t), y(t)$. This system uses $\mathbf{A} \in \mathbb{R}^{N \times N}$ as the evolution parameter and $\mathbf{B} \in \mathbb{R}^{N \times 1}, \mathbf{C} \in \mathbb{R}^{1 \times N}$ as the projection parameters.

Since SSMs operate on continuous sequences $x(t)$, they are unable to process discrete token inputs such as images and texts. Consequently, the adaptation to a discretized version of the SSM becomes essential. As described in Mamba~\cite{DBLP:journals/corr/abs-2312-00752}, the commonly used method for transformation $\mathbf{A},\mathbf{B}$ from continuous to discrete form is zero-order hold (ZOH), which is 
defined as follows:
\begin{equation}
\begin{split}
\overline{\mathbf{A}} &= \exp (\boldsymbol{\Delta} \mathbf{A}), \\
\overline{\mathbf{B}} &= (\boldsymbol{\Delta} \mathbf{A})^{-1}(\exp (\boldsymbol{\Delta} \mathbf{A})-\mathbf{I}) \cdot \boldsymbol{\Delta} \mathbf{B}.
\end{split}
\end{equation}
$\boldsymbol{\Delta}$ is the timescale parameter to transform the continuous parameters $\mathbf{A},\mathbf{B}$ to discrete parameters $\overline{\mathbf{A}},\overline{\mathbf{B}}$. After the operation, the discretized version of Eqn. \ref{first_defi} can be rewritten as:
\begin{equation}
\begin{split}
h_t & =\overline{\mathbf{A}} h_{t-1}+\overline{\mathbf{B}} x_t, \\
y_t & =\mathbf{C} h_t .
\end{split}
\end{equation}

At last, the models compute output through a global convolution:
\begin{equation}
\begin{split}
\overline{\mathbf{K}} & =\left(\mathbf{C} \overline{\mathbf{B}}, \mathbf{C} \overline{\mathbf{A B}}, \ldots, \mathbf{C} \overline{\mathbf{A}}^{L-1} \overline{\mathbf{B}}\right) \\
\mathbf{y} & =\mathbf{x} * \overline{\mathbf{K}},
\end{split}
\end{equation}
where $L$ is the length of the input sequence $\mathbf{x}$, $\overline{\mathbf{K}} \in \mathbb{R}^L$ is a structured convolutional kernel and $*$ represents the convolution operation.

\textbf{Selective SSMs.} The inherent LTI characteristic of SSMs, characterized by the consistent application of matrices $\overline{\mathbf{A}}$, $\overline{\mathbf{B}}$, $\overline{\mathbf{C}}$, and $\Delta$ across various inputs, limits their ability to adaptively filter and interpret contextual nuances within diverse input sequences. To overcome this limitation, we introduce Selective SSM as the core operator in our FractalMamba model. In Selective SSMs, the matrices $\overline{\mathbf{B}}$, $\overline{\mathbf{C}}$, and $\Delta$ are rendered as dynamic, input-responsive elements, effectively transitioning the SSM into a time-variant model. This modification enables the model to adapt more effectively to different input contexts, significantly enhancing its capacity to capture pertinent temporal features and relationships, and thereby improving the accuracy and efficiency of the input sequence representation.

\subsection{Limitations of Existing Scanning Mechanisms}
From Eqn. \ref{first_defi}, it becomes evident that selecting the most appropriate inputs for each time step in the SSM is crucial for effectively capturing and modeling the feature representation of the current image. To achieve this, the serialization process from 2D images to 1D sequences must meticulously capture the inherent structural information within the image. This requires preserving the structural coherence among image patches, ensuring that the serialized form maintains the essential spatial relationships present in the original image.

Existing linear scanning methods, such as the Z-order or Zigzag curve, exhibit significant limitations in preserving spatial relationships. While these methods maintain adjacency information between neighboring patches within the same row, they fail to effectively capture inter-row relationships. This oversight leads to the loss of crucial structural links essential for understanding broader spatial relationships within the image. Consequently, such scanning techniques compromise the structural integrity of the original image, undermining the model’s capacity to accurately reflect the image’s true characteristics. This lack of structural fidelity results in a model that is insensitive to variations in image scale, meaning that correlations evident at lower resolutions may not be preserved at higher resolutions. As a result, patterns and features learned at a lower resolution often do not translate effectively to higher resolutions, limiting the model’s applicability across different scales.

To overcome these challenges, we introduce fractal scanning mechanisms in our model. These mechanisms are designed to more adeptly follow the complex contours and structures within an image, thereby preserving its integral structural characteristics across different viewing scales. In this paper, we choose the Hilbert curve, which is a typical fractal curve.

\begin{wrapfigure}{r}{4cm}
\centering
\includegraphics[width=0.25\textwidth]{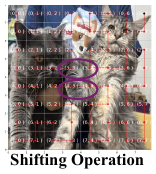}
\caption{The Hilbert curve and the corresponding shifting operation.}
\label{shift}
\end{wrapfigure}

\subsection{Fractal Scanning Mechanisms}
The Hilbert curve, a fractal defined through a recursive process, is particularly effective in image analysis because it maintains spatial and structural consistency at varying scales. Its property of self-similarity is vital for the analysis of high-resolution images, enabling it to capture features coherently across different scales. Starting from a single point, the Hilbert curve uses the midpoints of directional vectors \(\vec{x}\) and \(\vec{y}\) to dictate its path within the two-dimensional space. Through successive recursive subdivisions and traversals, the curve systematically accesses every part of the image, ensuring local continuity. This method effectively preserves the spatial relationships within the image, which is essential for delivering precise analysis and representation.

The Hilbert curve’s recursive design is inherently adaptable, allowing it to seamlessly conform to varying image resolutions. In this process, the vectors \(\vec{x}\) and \(\vec{y}\) are halved and swapped with each recursion, depending on the quadrant, illustrating the curve’s ability to efficiently and uniformly occupy two-dimensional spaces. This structured traversal technique ensures that image patches adjacent on the curve also remain adjacent in the image. Preserving this adjacency is key to maintaining the spatial relationships within the image. 

As shown in Algorithm. \ref{algorithm}, the use of vectors \(\vec{x}\) and \(\vec{y}\) in the pseudocode simplifies the understanding of directional movements and subdivisions within each recursive call, aligning the spatial mapping of the curve with the underlying pixel grid of the image. This alignment is pivotal in reducing the loss of important image features during the serialization processing, where maintaining the hierarchical structure of pixel data is paramount.

\begin{algorithm}[!t]
\caption{Hilbert Curve Generation}
\begin{algorithmic}[1]
\Function{Hilbert}{$x_0, y_0, \vec{x}, \vec{y}, depth$}
    \If{$depth = 0$}
        \State \textbf{plot}($x_0 + (\vec{x}[0] + \vec{y}[0])/2, y_0 + (\vec{x}[1] + \vec{y}[1])/2$)
    \Else
        \State \Call{Hilbert}{$x_0, y_0, \vec{y}/2, \vec{x}/2, depth - 1$}
        \State \Call{Hilbert}{$x_0 + \vec{x}[0]/2, y_0 + \vec{x}[1]/2, \vec{x}/2, \vec{y}/2, depth - 1$}
        \State \Call{Hilbert}{$x_0 + \vec{x}[0]/2 + \vec{y}[0]/2, y_0 + \vec{x}[1]/2 + \vec{y}[1]/2, \vec{x}/2, \vec{y}/2, depth - 1$}
        \State \Call{Hilbert}{$x_0 + \vec{x}[0]/2 + \vec{y}[0], y_0 + \vec{x}[1]/2 + \vec{y}[1], \vec{y}/2, \vec{x}/2, depth - 1$}
    \EndIf
\EndFunction
\label{pseudocode}
\end{algorithmic}
\label{algorithm}
\end{algorithm}

\textbf{Curves Shifting.} However, the use of fractal scanning curves in SSMs for image processing has revealed certain limitations, specifically concerning the imperfect maintenance of local adjacency and occasional disruptions in local continuity. As shown in Fig. \ref{shift}, the two patches in the image, (3,3) and (3,4), are neighboring in terms of spatial location, but the distance between the two of them is not neighboring in the Hilbert scan. These issues can lead to a partial loss of local proximity information, which is crucial for accurate modeling by SSMs. To address this challenge, we introduce a simple yet effective modification to the fractal curves. We apply a shift operation, moving the curves up or down by one pixel. This minor modification significantly improves the local adjacency and continuity of pixel serialization. By aligning the curve more closely with the inherent spatial relationships within the image, this shift helps to mitigate gaps and overlaps at critical junctions and complex parts of the curve, where adjacency might originally be lost. 

\section{Experiments} \label{Experiments}
In this section, we conduct a series of experiments to evaluate and compare FractalMamba with established benchmark models, including architectures based on CNNs and ViTs. Our assessment covers a range of visual tasks, such as image classification, object detection, and semantic segmentation. Following these evaluations, we provide a comprehensive analysis of FractalMamba’s characteristics, with a particular emphasis on its standout feature: the remarkable capability to efficiently handle increasingly large input resolutions. Note that, in the following image classification, object detection, and semantic segmentation evaluation, we do not add the shifting operation, since it would introduce extra curves in different directions. For fair comparison with other works lile VMamba, we only use four fractal curves. All experiment are conducted using 8 NVIDIA H800 GPUs.

\begin{table}[]
\centering
\caption{Performance comparison of our FractalMamba and other methods on ImageNet-1K.}
\begin{tabular}{@{}c|ccc|c@{}}
\toprule
Models         & Image Size & \#Param. & FLOPs & ImageNet Top-1 Acc. \\ \midrule
RegNetY-8G~\cite{DBLP:conf/cvpr/RadosavovicKGHD20}     & 224$^2$    & 39M      & 8.0G  & 81.7                \\
EffNet-B5~\cite{DBLP:conf/icml/TanL19}      & 456$^2$    & 30M      & 9.9G  & 83.6                \\
ViT-B/16~\cite{DBLP:conf/iclr/DosovitskiyB0WZ21}       & 384$^2$    & 86M      & 55.4G & 77.9                \\
DeiT-B~\cite{DBLP:conf/icml/TouvronCDMSJ21}         & 224$^2$    & 86M      & 17.5G & 81.8                \\
ConvNeXt-T~\cite{DBLP:conf/cvpr/0003MWFDX22}     & 224$^2$    & 29M      & 4.5G  & 82.1                \\
HiViT-S~\cite{DBLP:conf/iclr/0004TXHDY023}        & 224$^2$    & 38M      & 9.1G  & 83.5                \\
Swin-T~\cite{DBLP:conf/iccv/LiuL00W0LG21}         & 224$^2$    & 28M      & 4.6G  & 81.3                \\ \midrule
ViM-S~\cite{DBLP:journals/corr/abs-2401-09417}          & 224$^2$    & 26M      & -     & 80.5                \\
VMamba-T~\cite{DBLP:journals/corr/abs-2401-10166}       & 224$^2$    & 31M      & 4.9G  & 82.5                \\
LocalMamba-T~\cite{DBLP:journals/corr/abs-2403-09338}   & 224$^2$    & 26M      & 5.7G  & 82.7                \\
PlainMamba-L3~\cite{DBLP:journals/corr/abs-2403-17695}  & 224$^2$    & 50M      & 14.4G & 82.3                \\
\rowcolor[HTML]{C0C0C0} 
FractalMamba-T & 224$^2$    & 31M          & 4.9G      & 82.7                \\ \bottomrule
\end{tabular}
\label{Classification}
\end{table}

\subsection{Image Classification}
\textbf{Experiment Setting.} We evaluate the performance of FractalMamba using the ImageNet-1K dataset~\cite{deng2009imagenet}, following the evaluation protocol described in the work~\cite{liu2022swin}. The FractalMamba-T model is trained from scratch over 300 epochs, with an initial 20-epoch warm-up period, using a batch size of 1024. The training processes utilizes the AdamW optimizer~\cite{loshchilov2017decoupled}, with betas set to (0.9,0.999), momentum of 0.9, a cosine decay learning rate scheduler, an initial learning rate of $1 \times 10^{-3}$, and a weight decay of 0.05. Additional strategies such as label smoothing (0.1) and exponential moving average (EMA) were incorporated into the training regimen. No other training techniques were used beyond these specified methods.

\textbf{Model Performance.} The comparison results of FractalMamba against benchmark backbone models on the ImageNet-1K dataset are summarized in Table. \ref{Classification}. Notably, with comparable FLOPs, FractalMamba-T achieves a performance of 82.5\%, outperforming RegNetY-8G~\cite{DBLP:conf/cvpr/RadosavovicKGHD20} by 0.8\%, DeiT-S~\cite{DBLP:conf/icml/TouvronCDMSJ21} by 0.7\%, and Swin-T by 1.2\%. Moreover, FractalMamba demonstrates competitive performance relative to other SSM-based models. One primary reason is that for images with low and fixed resolutions, linear scanning suffices for the model to learn the corresponding patterns, and thus the fractal scanning mechanism does not exhibit a distinct performance advantage. However, as detailed in Section \ref{Analysis}, fractal curves, due to their unique self-similarity, effectively capture the structural information of images across varying resolutions, thereby exhibiting superior performance.

\begin{table}[]
\centering
\caption{The results of object detection and instance segmentation on the COCO dataset. FLOPs are calculated for an input size of $1280 \times 800$. The metrics $AP^b$ and $AP^m$ represent box AP and mask AP, respectively. The notation `1 $\times$' indicates that models were fine-tuned for 12 epochs, while `3 $\times$ MS’ denotes the utilization of multi-scale training across 36 epochs.}
\begin{tabular}{@{}ccccccccc@{}}
\toprule
\multicolumn{9}{c}{\textbf{Mask R-CNN 1 $\times$ schedule}} \\ \midrule
\multicolumn{1}{c|}{Backbone} &
  AP$^b$ &
  AP$^b_{50}$ &
  \multicolumn{1}{c|}{AP$^b_{75}$} &
  AP$^m$ &
  AP$^m_{50}$ &
  \multicolumn{1}{c|}{AP$^m_{75}$} &
  \#param. &
  FLOPs \\ \midrule
\multicolumn{1}{c|}{Swin-T} &
  42.7 &
  65.2 &
  \multicolumn{1}{c|}{46.8} &
  39.3 &
  62.2 &
  \multicolumn{1}{c|}{42.2} &
  48M &
  267G \\
\multicolumn{1}{c|}{ConvNeXt-T} &
  44.2 &
  66.6 &
  \multicolumn{1}{c|}{48.3} &
  40.1 &
  63.3 &
  \multicolumn{1}{c|}{42.8} &
  48M &
  262G \\
\multicolumn{1}{c|}{ViT-Adapter-S} &
  44.7 &
  65.8 &
  \multicolumn{1}{c|}{48.3} &
  39.9 &
  62.5 &
  \multicolumn{1}{c|}{42.8} &
  48M &
  403G \\
\multicolumn{1}{c|}{VMamba-T} &
  47.4 &
  69.5 &
  \multicolumn{1}{c|}{52.0} &
  42.7 &
  66.3 &
  \multicolumn{1}{c|}{46.0} &
  50M &
  270G \\
\multicolumn{1}{c|}{LocalMamba-T} &
  46.7 &
  68.7 &
  \multicolumn{1}{c|}{50.8} &
  42.2 &
  65.7 &
  \multicolumn{1}{c|}{45.5} &
  45M &
  291G \\
\multicolumn{1}{c|}{PlainMamba-L3} &
  46.8 &
  68.0 &
  \multicolumn{1}{c|}{51.1} &
  41.2 &
  64.7 &
  \multicolumn{1}{c|}{43.9} &
  79M &
  696G \\
\rowcolor[HTML]{C0C0C0} 
\multicolumn{1}{c|}{\cellcolor[HTML]{C0C0C0}FractalMamba-T} &
  47.8 &
  70.0 &
  \multicolumn{1}{c|}{\cellcolor[HTML]{C0C0C0}52.4} &
  42.9 &
  66.6 &
  \multicolumn{1}{c|}{\cellcolor[HTML]{C0C0C0}46.3} &
  50M &
  270G \\ \midrule
\multicolumn{9}{c}{\textbf{Mask R-CNN 3 $\times$ MS schedule}} \\ \midrule
\multicolumn{1}{c|}{Swin-T} &
  46.0 &
  68.1 &
  \multicolumn{1}{c|}{50.3} &
  41.6 &
  65.1 &
  \multicolumn{1}{c|}{44.9} &
  48M &
  267G \\
\multicolumn{1}{c|}{ConvNeXt-T} &
  46.2 &
  67.9 &
  \multicolumn{1}{c|}{50.8} &
  41.7 &
  65.0 &
  \multicolumn{1}{c|}{44.9} &
  48M &
  262G \\
\multicolumn{1}{c|}{ViT-Adapter-S} &
  48.2 &
  69.7 &
  \multicolumn{1}{c|}{52.5} &
  42.8 &
  66.4 &
  \multicolumn{1}{c|}{45.9} &
  48M &
  403G \\
\multicolumn{1}{c|}{VMamba-T} &
  48.9 &
  70.6 &
  \multicolumn{1}{c|}{53.6} &
  43.7 &
  67.7 &
  \multicolumn{1}{c|}{46.8} &
  50M &
  270G \\
\multicolumn{1}{c|}{LocalMamba-T} &
  48.7 &
  70.1 &
  \multicolumn{1}{c|}{53.0} &
  43.4 &
  67.0 &
  \multicolumn{1}{c|}{46.4} &
  45M &
  291G \\
\rowcolor[HTML]{C0C0C0} 
\multicolumn{1}{c|}{\cellcolor[HTML]{C0C0C0}FractalMamba-T} &
  49.5 &
  71.3 &
  \multicolumn{1}{c|}{\cellcolor[HTML]{C0C0C0}54.0} &
  44.1 &
  68.5 &
  \multicolumn{1}{c|}{\cellcolor[HTML]{C0C0C0}47.4} &
  50M &
  270G \\ \bottomrule
\end{tabular}
\label{Detection}
\end{table}

\subsection{Object Detection}
\textbf{Experiment Setting.} In this section, we evaluate the performance of FractalMamba on object detection using the MSCOCO 2017 dataset~\cite{lin2014microsoft}. We configure our training setup using the MMDetection library~\cite{chen2019mmdetection}, adhering to the hyper-parameters utilized in the Swin~\cite{DBLP:conf/iccv/LiuL00W0LG21} model with the Mask-RCNN detector. Specifically, we use the AdamW optimizer~\cite{loshchilov2017decoupled} and fine-tune the pre-trained classification models (originally trained on ImageNet-1K) over both 12 and 36 epochs. For FractalMamba-T, the drop path rate is set at 0.2\%. The learning rate starts at $1 \times 10^{-4}$ and is reduced by a factor of 10 at the 9th and 11th epochs. Multi-scale training and random flipping are implemented with a batch size of 16, aligning with established best practices for object detection evaluations.

\textbf{Model Performance.} The results on the COCO dataset are summarized in Table \ref{Detection}. FractalMamba consistently outperforms other models in both box and mask AP, regardless of the training schedule employed. Specifically, with a 12-epoch fine-tuning schedule, FractalMamba-T models achieve object detection mean Average Precision (mAP) of 47.8\%, which is superior to Swin-T by 5.1\% mAP, ConvNeXt-T by 3.6\% mAP and VMamba-T by 0.4\%. In the same configuration, FractalMamba-T achieves instance segmentation mean Intersection over Union (mIoU) of 42.9\%, surpassing Swin-T by 3.6\% mIoU and ConvNeXt-T by 2.7\% mIoU. These results highlight FractalMamba’s capability to deliver robust performance in downstream tasks requiring dense prediction.

\begin{table}[]
\centering
\caption{Results of semantic segmentation on ADE20K using UperNet~\cite{xiao2018unified}. FLOPs are calculated with input size of $512 \times 2048$. `SS' and `MS' denote single-scale and multi-scale testing, respectively.}
\begin{tabular}{@{}c|c|cc|cc@{}}
\toprule
Models         & Crop Size & mIoU(SS) & mIoU(MS) & \#param. & FLOPs \\ \midrule
ResNet-50      & 512$^2$   & 42.1     & 42.8     & 67M      & 953G  \\
DeiT-S+MLN     & 512$^2$   & 43.8     & 45.1     & 58M      & 1217G \\
Swin-T         & 512$^2$   & 44.4     & 45.8     & 60M      & 945G  \\
ConvNeXt-T     & 512$^2$   & 46.0     & 46.7     & 60M      & 939G  \\
VMamba-T       & 512$^2$   & 48.3     & 48.6     & 62M      & 948G  \\
LocalMamba-T   & 512$^2$   & 47.9     & 49.1     & 57M      & 970G  \\
PlainMamba-L2  & 512$^2$   & 46.8     & -        & 55M      & 285G  \\
\rowcolor[HTML]{C0C0C0} 
FractalMamba-T & 512$^2$   & 48.9     & 49.8     & 62M      & 948G      \\ \bottomrule
\end{tabular}
\label{Segmentation}
\end{table}

\subsection{Semantic Segmentation}
\textbf{Experiment Setting.} Following Swin~\cite{DBLP:conf/iccv/LiuL00W0LG21}, we augment the pre-trained model with an UperHead~\cite{xiao2018unified}. We employ the AdamW optimizer~\cite{loshchilov2017decoupled}, setting the learning rate to $6 \times 10^{-5}$. The fine-tuning process extends over 160,000 iterations with a batch size of 16. The standard input resolution is $512 \times 512$, and we additionally provide experimental results using $640 \times 640$ inputs along with multi-scale (MS) testing to evaluate performance enhancements at varied resolutions.

\textbf{Model Performance.} The results of semantic segmentation on the ADE20K dataset are summarized in Table. \ref{Segmentation}. In line with findings from previous experiments, FractalMamba exhibits superior accuracy. Specifically, FractalMamba-T achieves a mean Intersection over Union (mIoU) of 48.9\% with a resolution of $512 \times 512$, and 49.8\% mIoU with multiscale (MS) input. These results surpass those of all benchmarked methods, including ResNet~\cite{DBLP:conf/cvpr/HeZRS16}, DeiT~\cite{DBLP:conf/icml/TouvronCDMSJ21}, Swin~\cite{DBLP:conf/iccv/LiuL00W0LG21}, and ConvNeXt~\cite{DBLP:conf/cvpr/0003MWFDX22}, confirming FractalMamba’s effectiveness in semantic segmentation tasks.

\begin{table}[]
\centering
\caption{The effectiveness of our proposed shifting operation.}
\begin{tabular}{@{}c|cccccc@{}}
\toprule
Models                & 224$^2$ & 384$^2$ & 512$^2$ & 640$^2$ & 768$^2$ & 1024$^2$ \\ \midrule
FractalMamba          & 82.7    & 82.4    & 81.2    & 80.2    & 77.9    & 69.6     \\
FractalMamba+Shifting & 82.9    & 82.7    & 81.6    & 80.5    & 78.3    & 70.4     \\ \bottomrule
\end{tabular}
\label{abla_Shifting}
\end{table}

\begin{table}[]
\centering
\caption{Comparison of generalizability to image inputs with different spatial resolutions.}
\begin{tabular}{@{}c|ccc|c@{}}
\toprule
Models                                          & Image Size & \#Param. & FLOPs  & ImageNet Top-1 Acc. \\ \midrule
                                                & 224$^2$    & 26M      & 4.1G   & 76.4                \\
                                                & 384$^2$    & 26M      & 12.1G  & 76.5                \\
                                                & 512$^2$    & 26M      & 21.5G  & 73.4                \\
                                                & 640$^2$    & 26M      & 33.5G  & 69.7                \\
                                                & 768$^2$    & 26M      & 48.3G  & 65.3                \\
\multirow{-6}{*}{ResNet-50}                     & 1024$^2$   & 26M      & 85.9G  & 52.1                \\ \midrule
                                                & 224$^2$    & 29M      & 4.5G   & 82.0                \\
                                                & 384$^2$    & 29M      & 13.1G  & 81.0                \\
                                                & 512$^2$    & 29M      & 23.3G  & 78.0                \\
                                                & 640$^2$    & 29M      & 36.5G  & 74.3                \\
                                                & 768$^2$    & 29M      & 52.5G  & 69.5                \\
\multirow{-6}{*}{ConvNeXt}                      & 1024$^2$   & 29M      & 93.3G  & 55.4                \\ \midrule
                                                & 224$^2$    & 22M      & 4.6G   & 80.7                \\
                                                & 384$^2$    & 22M      & 15.5G  & 78.9                \\
                                                & 512$^2$    & 22M      & 31.8G  & 74.2                \\
                                                & 640$^2$    & 22M      & 58.2G  & 68.0                \\
                                                & 768$^2$    & 22M      & 98.7G  & 70.0                \\
\multirow{-6}{*}{DeiT}                          & 1024$^2$   & 22M      & 243.1G & 46.9                \\ \midrule
                                                & 224$^2$    & 28M      & 4.5G   & 81.2                \\
                                                & 384$^2$    & 28M      & 14.5G  & 80.7                \\
                                                & 512$^2$    & 28M      & 26.6G  & 79.0                \\
                                                & 640$^2$    & 28M      & 45.0G  & 76.6                \\
                                                & 768$^2$    & 28M      & 70.7G  & 73.1                \\
\multirow{-6}{*}{Swin}                          & 1024$^2$   & 28M      & 152.5G & 61.9                \\ \midrule
                                                & 224$^2$    & 19M      & 4.6G   & 81.9                \\
                                                & 384$^2$    & 19M      & 15.2G  & 81.5                \\
                                                & 512$^2$    & 19M      & 30.6G  & 79.3                \\
                                                & 640$^2$    & 19M      & 54.8G  & 76.0                \\
                                                & 768$^2$    & 19M      & 91.4G  & 71.4                \\
\multirow{-6}{*}{HiViT}                         & 1024$^2$   & 19M      & 218.9G & 58.9                \\ \midrule
                                                & 224$^2$    & 31M      & 4.9G   & 82.5                \\
                                                & 384$^2$    & 31M      & 14.3G  & 82.5                \\
                                                & 512$^2$    & 31M      & 25.4G  & 81.1                \\
                                                & 640$^2$    & 31M      & 39.6G  & 79.3                \\
                                                & 768$^2$    & 31M      & 57.1G  & 76.1                \\
\multirow{-6}{*}{VMamba}                        & 1024$^2$   & 31M      & 101.5G & 62.3                \\ \midrule
\cellcolor[HTML]{FFFFFF}{\color[HTML]{333333} } & 224$^2$    & 31M      & 4.9G      & 82.5                \\
\cellcolor[HTML]{FFFFFF}{\color[HTML]{333333} } & 384$^2$    & 31M      & 14.3G       & 82.4                \\
\cellcolor[HTML]{FFFFFF}{\color[HTML]{333333} } & 512$^2$    & 31M      & 25.4G       & 81.2                \\
\cellcolor[HTML]{FFFFFF}{\color[HTML]{333333} } & 640$^2$    & 31M      & 39.6G       &  80.2                   \\
\cellcolor[HTML]{FFFFFF}{\color[HTML]{333333} } & 768$^2$    & 31M      & 57.1G       & 77.9                \\
\multirow{-6}{*}{\cellcolor[HTML]{FFFFFF}{\color[HTML]{333333} FractalMamba}} & 1024$^2$ & 31M &101.5G  & 69.6 \\ \bottomrule
\end{tabular}
\label{diff_scales}
\end{table}

\subsection{Ablation Studies} \label{Analysis}

\textbf{The effectiveness of Shifting Operation.} We identify an issue where the inherent structure of fractal curves could lead to a partial loss of local proximity information—an essential element for accurate modeling by SSMs. To overcome this limitation, we introduced a straightforward yet effective modification to the fractal curves by implementing a shift operation. This adjustment involves shifting the entire curve vertically or horizontally by one pixel. The effectiveness of this approach is quantitatively evaluated in Table. \ref{abla_Shifting}, where the adjusted fractal curves consistently outperform the traditional configurations in maintaining higher fidelity in feature representation.
\begin{figure}[!t]
    \centering
    \includegraphics[width=\linewidth]{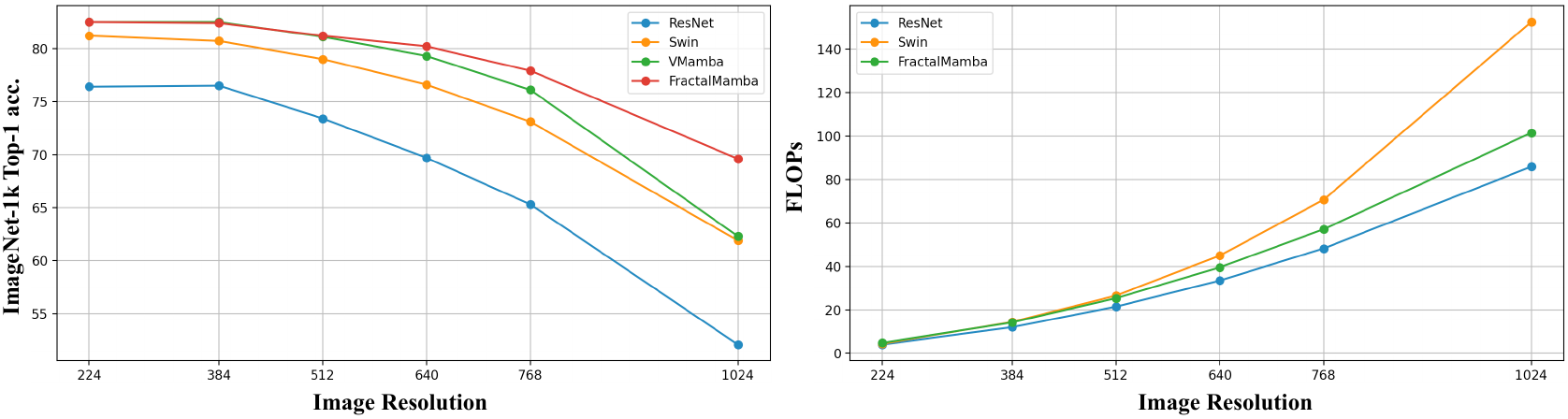}
    \caption{The variations in classification accuracy and computational complexity (FLOPs) as the resolutions of test images increase progressively.}
    \label{diff_scales}
\end{figure}

\textbf{Model Scalability and Efficiency.} In this study, we conduct experiments to evaluate the scalability of FractalMamba and other SSM-based models with progressively larger input images. Specifically, models initially trained with an input size of $224 \times 224$ are applied directly to images with resolutions ranging from $384 \times 384$ to $1024 \times 1024$. We assess their generalization performance by examining the number of parameters, FLOPs during both training and inference phases, and Top-1 classification accuracy on the ImageNet-1K dataset.

According to the results summarized in Table \ref{diff_scales}, FractalMamba exhibits the most consistent performance across varying input sizes, achieving a Top-1 classification accuracy of 69.6\% with 101.5G FLOPs when the input resolution is increased to $1024 \times 1024$. In comparison, Swin~\cite{DBLP:conf/iccv/LiuL00W0LG21}, with the same input size, achieves a Top-1 accuracy of 61.9\% but experiences a significant drop to 19. In contrast, while ResNet50~\cite{DBLP:conf/cvpr/HeZRS16} maintains a relatively high inference speed at the largest input resolution, its classification accuracy drops to 52.1\%, underscoring its limited scalability. The performance changes across various resolutions for different models are intuitively depicted in Fig. \ref{diff_scales}. Notably, FractalMamba demonstrates a linear increase in computational complexity, as measured by FLOPs, closely mirroring the behavior of traditional CNN-based architectures. This observation substantiates the theoretical predictions regarding selective SSMs~\cite{DBLP:journals/corr/abs-2312-00752}.

\section{Conclusion}
In this paper, we introduce a novel method for serializing image patches using fractal scanning curves to enhance the performance of SSMs in various computer vision tasks. Unlike traditional linear scanning curves, fractal curves exhibit superior handling of images across multiple scales by maintaining high spatial proximity and adapting seamlessly to different image resolutions. This approach not only reduces redundancy but also more accurately captures complex patterns within images. We validate our method across a range of computer vision tasks, including image classification, detection, and segmentation. The experimental results unequivocally demonstrate that fractal curve scanning significantly outperforms linear curve scanning in these applications. These findings underscore the practicality of fractal curves in vision tasks and pave the way for future research, such as exploring additional fractal scanning methods to further enhance model performance. We believe that with continued refinement, fractal curve usage in SSMs will become increasingly pivotal in future computer vision applications, particularly in processing high-resolution and large-scale image data. By advancing these models, our goal is to develop more efficient and accurate vision processing technologies to address the escalating demands of image data processing.

{\small
\bibliography{reference}
\bibliographystyle{ieee_fullname}
}

\end{document}